\documentclass[10pt,twocolumn,letterpaper]{article}

\usepackage{cvpr}              

\usepackage{graphicx}
\usepackage{amsmath}
\usepackage{amssymb}
\usepackage{booktabs}
\usepackage{algorithm}
\usepackage{algpseudocode}
\usepackage{caption}
\usepackage{subcaption}

\usepackage{comment}

\usepackage{ifthen}
\newboolean{ack}
\setboolean{ack}{true}

\usepackage[pagebackref,breaklinks,colorlinks]{hyperref}

\usepackage[capitalize]{cleveref}
\crefname{section}{Sec.}{Secs.}
\Crefname{section}{Section}{Sections}
\Crefname{table}{Table}{Tables}
\crefname{table}{Tab.}{Tabs.}

\def\cvprPaperID{7} 
\def\confName{CVPR}
\def\confYear{2023}

\begin{document}

\title{How Efficient Are Today's Continual Learning Algorithms?}

\author{Md~Yousuf~Harun$^{1}$ \qquad Jhair~Gallardo$^{1}$ \qquad Tyler~L.~Hayes$^1$\thanks{Now at NAVER LABS Europe.} \qquad Christopher~Kanan$^{2}$\\
Rochester Institute of Technology$^1$, University of Rochester$^2$\\
{\tt\small \{mh1023, gg4099, tlh6792\}@rit.edu, }{\tt\small ckanan@cs.rochester.edu}\\
}

\maketitle

\begin{abstract}
   Supervised Continual learning involves updating a deep neural network (DNN) from an ever-growing stream of labeled data. While most work has focused on overcoming catastrophic forgetting, one of the major motivations behind continual learning is being able to efficiently update a network with new information, rather than retraining from scratch on the training dataset as it grows over time. Despite recent continual learning methods largely solving the catastrophic forgetting problem, there has been little attention paid to the efficiency of these algorithms. Here, we study recent methods for incremental class learning and illustrate that many are highly inefficient in terms of compute, memory, and storage. Some methods even require more compute than training from scratch! We argue that for continual learning to have real-world applicability, the research community cannot ignore the resources used by these algorithms. There is more to continual learning than mitigating catastrophic forgetting.
\end{abstract}

\section{Introduction}
\label{sec:intro}
In many real-world scenarios, the dataset used to train a deep neural network (DNN) grows over time. In industry settings, this is typically handled by periodically re-training the DNN from scratch after the dataset has grown (i.e., offline training); however, this is highly sub-optimal. In contrast, continual learning (CL) systems have the ability to update on novel inputs over time~\cite{parisi2019continual}. CL has the potential to yield significant computational benefits over periodically re-training from scratch. The vast majority of CL research has focused on solving catastrophic forgetting, which occurs when updating a DNN only on new data with non-CL methods. Thus, many CL algorithms have been designed for \emph{class incremental learning}, since it is a problem where severe catastrophic forgetting occurs~\cite{kemker2018measuring}. In this setting, the learner incrementally learns batches of mutually exclusive subsets of classes, rather than learning them all at once as in the offline setting. While state-of-the-art algorithms, e.g., DyTox~\cite{douillard2022dytox} and DER~\cite{yan2021dynamically}, excel at class incremental learning under a limited replay memory budget, they ignore other critical factors such as model size, compute, total memory usage, data efficiency, and training time. We argue that these factors cannot be ignored. Here, we study the efficiency of existing CL methods for class incremental learning on ImageNet, which reveals that many of them are impractical for real-world CL applications.

\begin{figure}[t]
  \centering
  \vspace{-1em}
   \includegraphics[width=\linewidth]{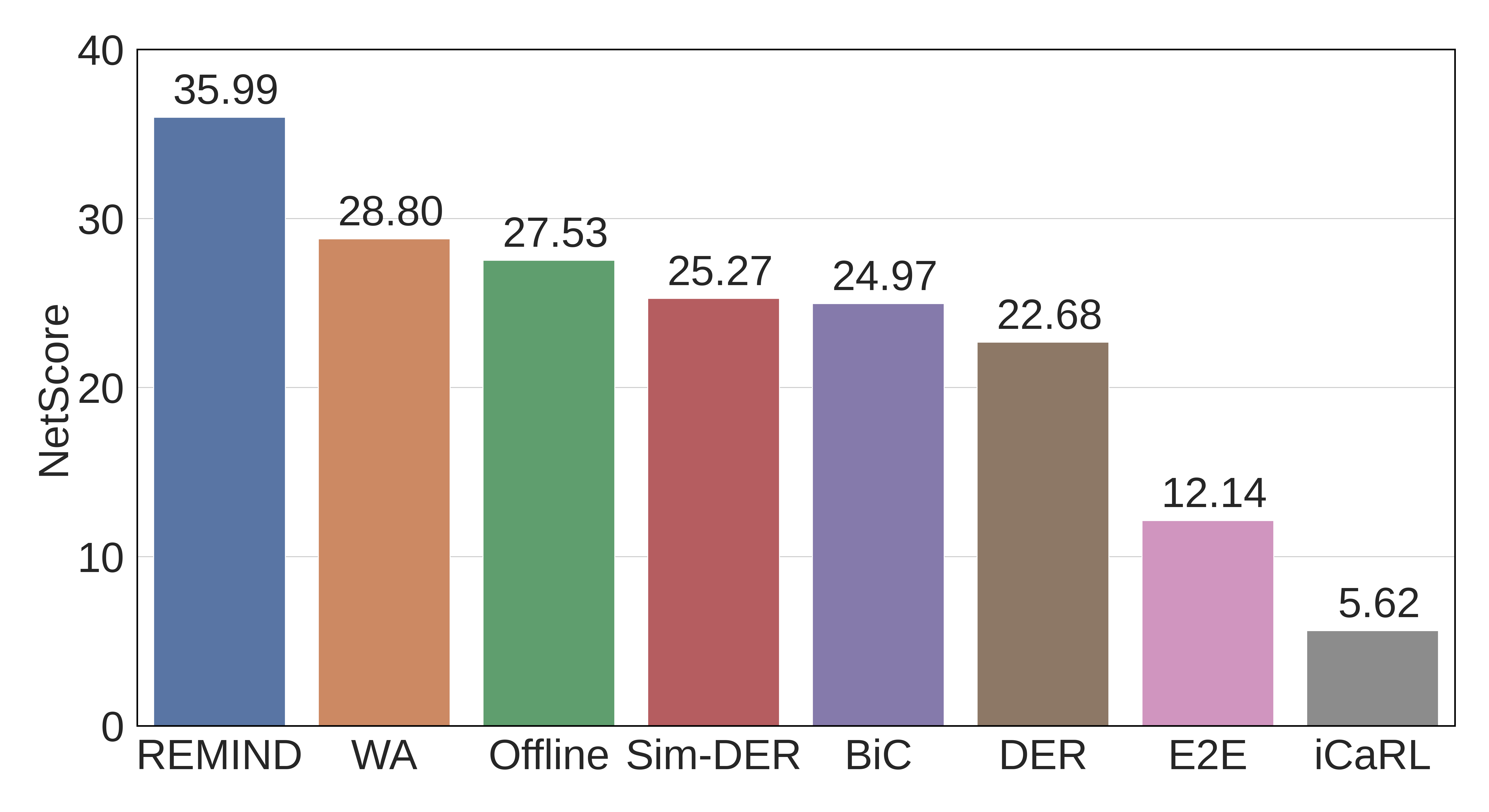}
   \caption{We use NetScore to evaluate CL methods based on their accuracy, parameter count, amount of memory used, and number of backpropagation updates (compute).
   }
   \label{fig:cost_acc}
\end{figure}

Most CL methods for incremental class learning on ImageNet perform a very large number of updates to the DNN, where we define an update as backpropagation on a single input. In opposition to one of the major goals of CL, computational efficiency, some models have become \emph{more} computationally expensive than an offline model trained on all data. For example, on ImageNet, DER was used to incrementally train a ResNet18 model, but this required more updates ($213.17$ Million) than training ResNet18 offline ($115.31$ Million). They also use many more parameters than the offline DNN, e.g., DER~\cite{yan2021dynamically} combines multiple feature extraction networks (i.e., one network per task for each 100 class increment) to learn a unified classifier that expands the total number of parameters to $10\times$ the offline DNN.

Another shortcoming of existing systems is that they do not fully account for the amount of memory or storage used during incremental training. State-of-the-art systems for incremental class learning on ImageNet use replay to mitigate catastrophic forgetting, where past data, which is stored in a limited buffer, is mixed (i.e., replayed) with new incoming data. However, many of them, including iCaRL~\cite{rebuffi2017icarl}, End-to-End~\cite{castro2018end}, BiC~\cite{wu2019large}, WA~\cite{zhao2020maintaining}, and DER~\cite{yan2021dynamically}, 
also keep new data temporarily in memory (or disk storage) during each batch, which is unaccounted for. Each incremental learning batch typically consists of around 100,000 images (100 classes). While smaller batches could be used to reduce memory overhead, it has been shown that many methods suffer from severe catastrophic forgetting unless large batches are used~\cite{hayes2020REMIND}. Additional memory is also used for the DNN parameters.

Ideally, a model should adapt to a growing training dataset without increasing the computational or memory budget. However, most CL methods lack this ability. While memory and storage may not be a concern for training in industry settings, it does matter for on-device learning applications, including autonomous robots, smart appliances, mobile phones, virtual/ augmented reality (VR/AR) headsets, and other wearable devices, which typically have very limited storage. For on-device learning, a CL system must adapt to new information quickly despite limited computation and memory.

\paragraph{Our contributions can be summarized as follows:}
\begin{enumerate}
\item We use the NetScore metric~\cite{hayes2022online} for evaluating state-of-the-art methods for incremental class learning on ImageNet-1K to summarize their properties in terms of accuracy, memory, parameters, and compute needed for training.
\item We find that most methods are grossly inefficient. Multiple methods use more compute than an offline learner, defeating one of the major reasons for studying CL.
\item We discuss the criteria needed for CL to be useful for real-world applications, and we call for the CL research community to focus on issues beyond catastrophic forgetting.
\end{enumerate}

\section{Methodology}
\label{sec:method}

\subsection{Evaluation Criteria}

We evaluate a model's efficiency in terms of four criteria: accuracy, model size, computational overhead (updates), and memory overhead. We adopt a modified variant of the \emph{NetScore} metric from \cite{hayes2022online} that combines these criteria. For a model $\mathcal{G}$, the \emph{NetScore} metric $\Omega(\mathcal{G})$ is defined as:
\begin{equation}
\Omega(\mathcal{G}) = s \log{(\frac{a(\mathcal{G})^{\alpha}}{p(\mathcal{G})^{\beta} u(\mathcal{G})^{\gamma} m(\mathcal{G})^{\zeta}})} \enspace ,
\end{equation}
where $a(\mathcal{G})$ is the final top-5 accuracy, $p(\mathcal{G})$ is number of parameters, $u(\mathcal{G})$ is the number of backpropagation updates, and $m(\mathcal{G})$ is the total memory usage.
The coefficients $\alpha$, $\beta$, $\gamma$, and $\zeta$ control the contribution of each factor.  Following \cite{hayes2022online}, we set $\alpha=2$ and $s=20$. We set
$\beta=\gamma=\zeta=0.125$ to avoid having a negative $\Omega(\mathcal{G})$.

\subsection{Algorithms Studied}

We evaluate the following CL methods, which have all been shown to perform well on ImageNet-1K:
\begin{itemize}
    \item iCaRL~\cite{rebuffi2017icarl} learns representations using a distillation loss and makes predictions using a nearest class mean classifier in feature space. During each incremental learning step, iCaRL trains the entire network.
    \item End-to-End~\cite{castro2018end} upgrades iCaRL using augmentations and fine-tunes the CNN's output layer on a balanced set rather than using a nearest class mean classifier. The augmentations applied in End-to-End include brightness enhancement, contrast normalization, random crops, and mirror flips.
    \item BiC~\cite{wu2019large} upgrades iCaRL by re-adjusting the logits of new classes by training a linear model on a validation set. For this, two bias correction parameters on the final layer are optimized.
    \item WA~\cite{zhao2020maintaining} upgrades iCaRL by aligning the norms of new class parameters to those of old class parameters.
    \item DER~\cite{yan2021dynamically} augments previously learned representations with representations for the new classes. It has a separate feature extractor per incremental batch 
    and applies pruning to reduce number of parameters. It also employs an auxiliary classifier to discriminate between past and present observations. We compare three variants of DER: DER without pruning, DER with pruning, and Simple-DER~\cite{li2021preserving}.
    \item  REMIND~\cite{hayes2020REMIND} uses compressed feature replay where mid-level CNN features are quantized to reduce memory overhead. In contrast, the other methods use replay of stored images. REMIND freezes earlier CNN layers and trains the remaining layers on reconstructed features. To do this, REMIND uses the first batch to initialize the CNN. 
\end{itemize}

\begin{figure*}[t]
  \centering

\begin{subfigure}[b]{0.3\textwidth}
         \centering
         \includegraphics[width=\textwidth]{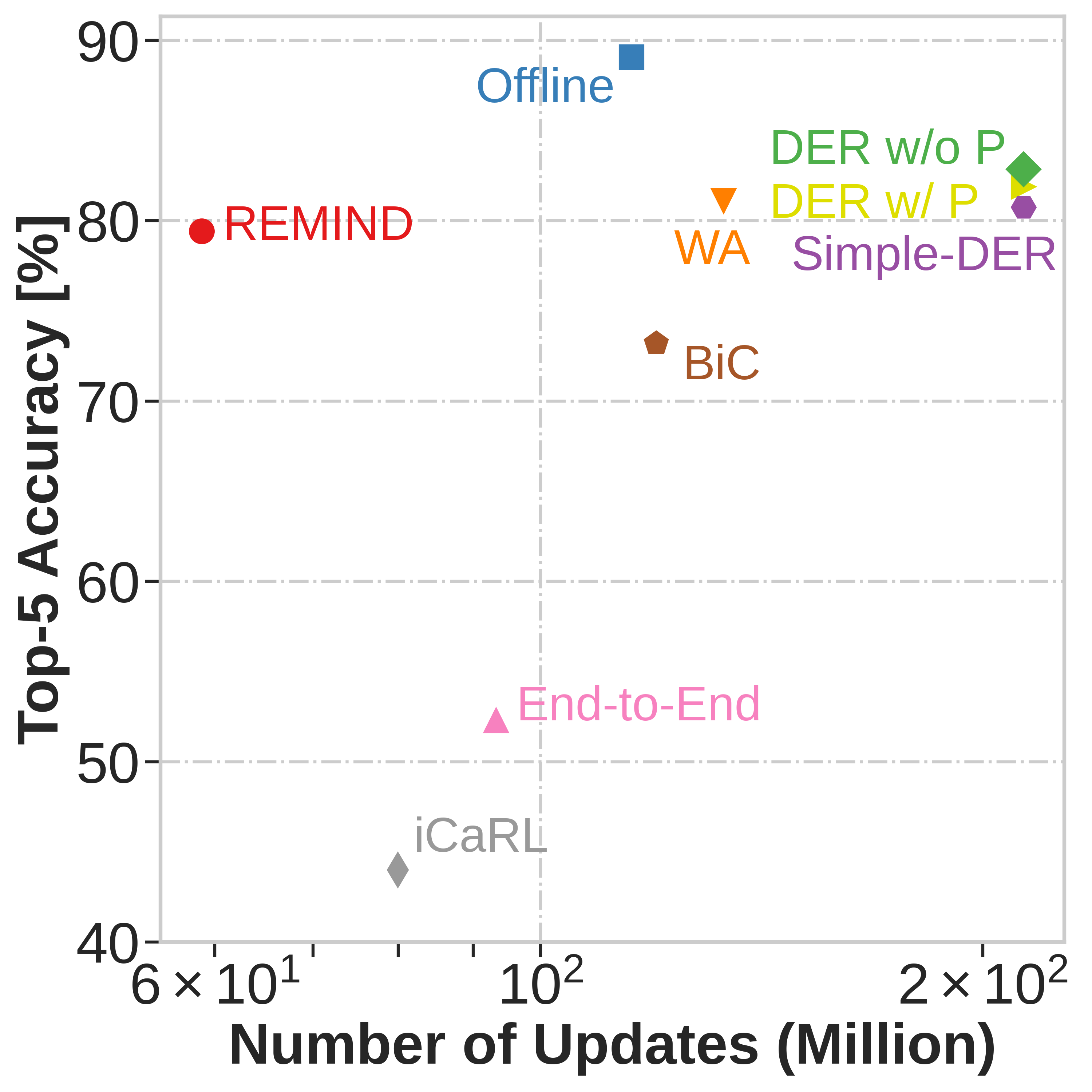}
         \caption{Accuracy vs. Backprop Updates}
         \label{fig:updates}
     \end{subfigure}
     \hfill
     \begin{subfigure}[b]{0.3\textwidth}
         \centering
         \includegraphics[width=\textwidth]{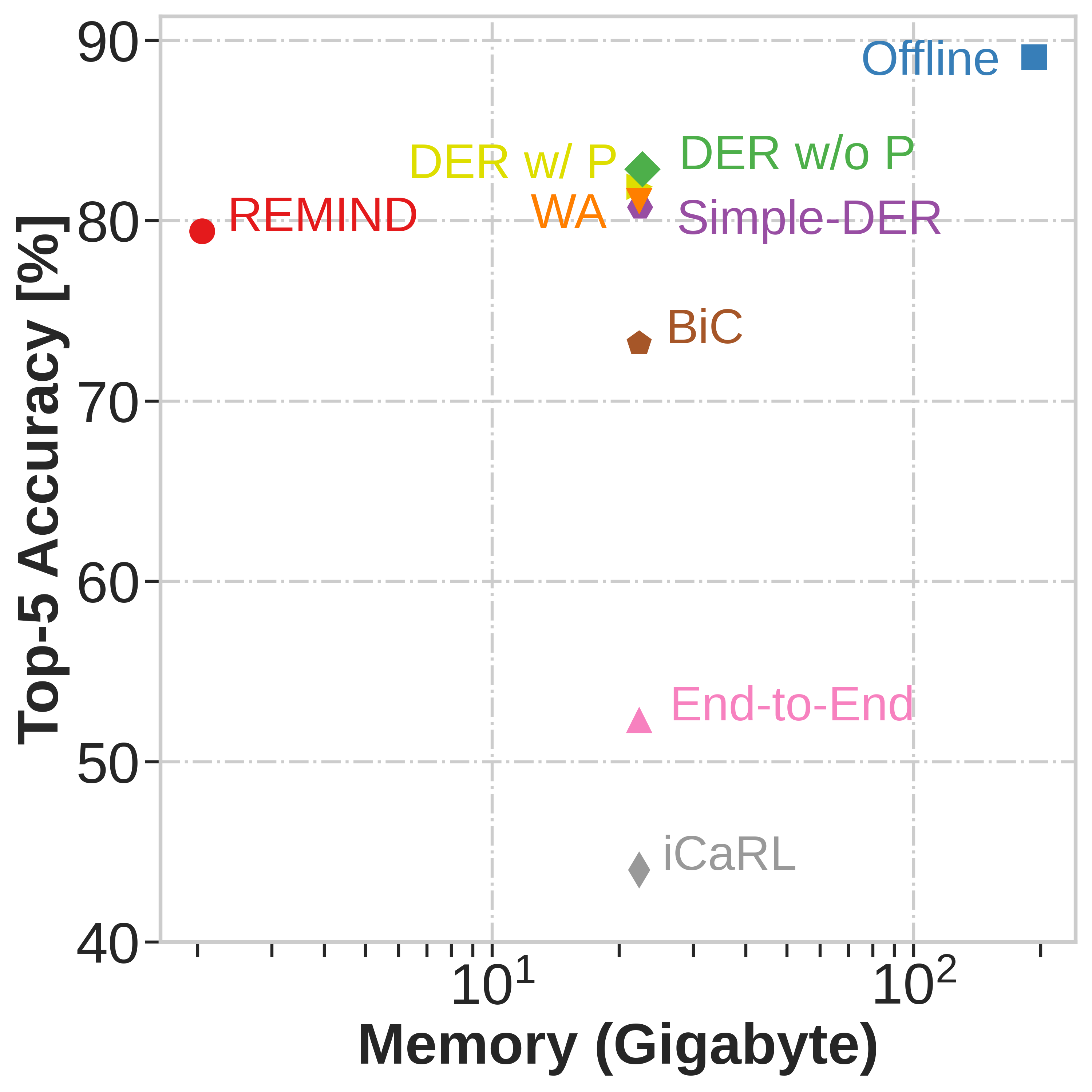}
         \caption{Accuracy vs. Memory}
         \label{fig:memory}
     \end{subfigure}
     \hfill
     \begin{subfigure}[b]{0.3\textwidth}
         \centering
         \includegraphics[width=\textwidth]{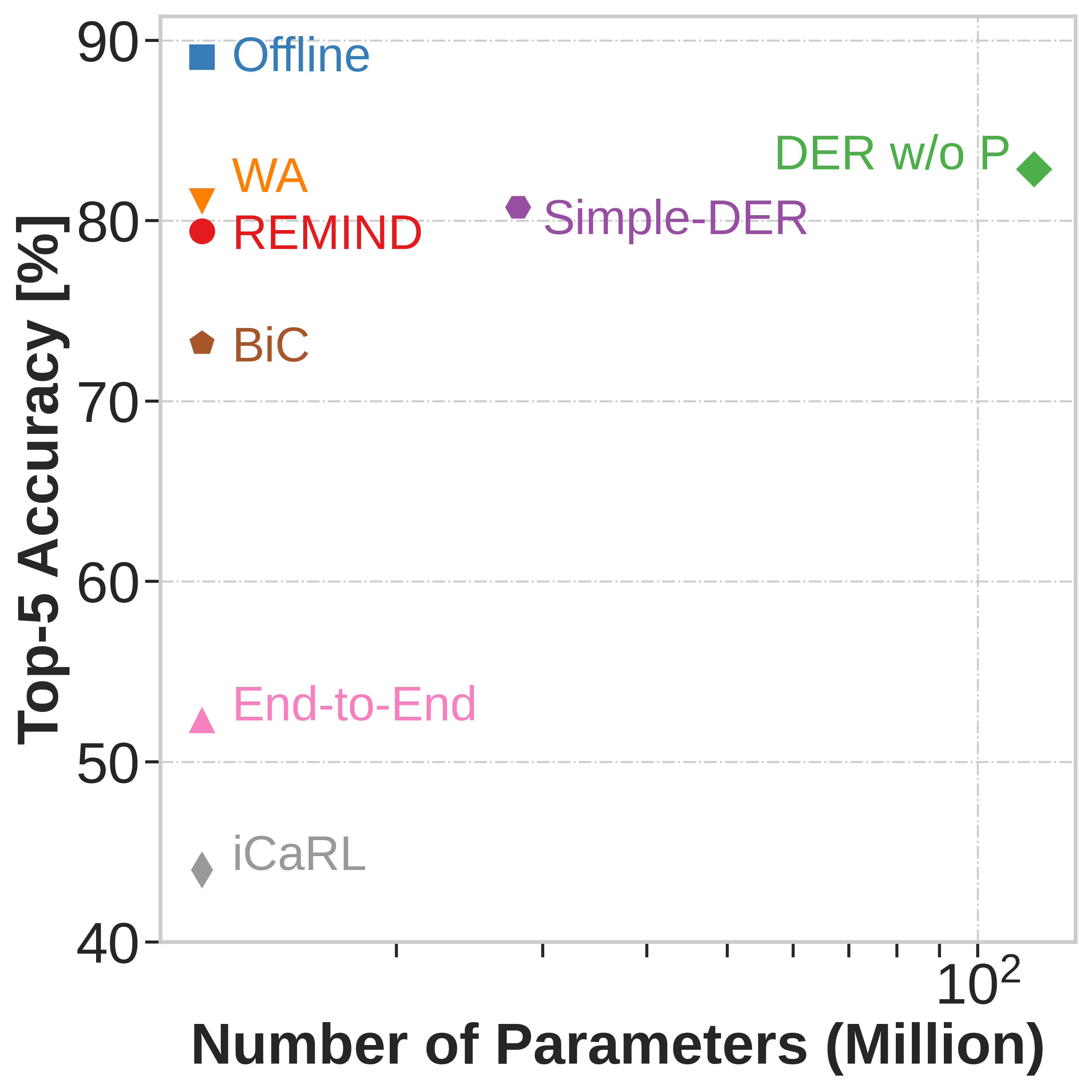}
         \caption{Accuracy vs. Parameter Count}
         \label{fig:parameters}
     \end{subfigure}       
   \caption{Comparison of methods in terms of top-5 accuracy and backprop updates, memory, and parameter count. In all sub-figures, REMIND uses the fewest resources and provides competitive accuracy. In contrast, DER w/o P uses the most resources, but achieves the highest accuracy compared to other CL methods. In (a), several methods such as BiC, WA, DER w/o P, DER w/ P, and Simple-DER require more computation than offline ResNet18. In (c), DER w/ P is not included since it does not report total number of parameters.
   \label{fig:stratified-perf}}
\end{figure*}

For our analysis, except REMIND, we use the numbers 
as reported in each method's paper on ImageNet and the settings they describe in their work. Since the data ordering used in the REMIND paper~\cite{hayes2020REMIND} is different, we implement REMIND with the same data ordering used by the compared methods in Table~\ref{tab:inc}. For replay,
REMIND uses 2 GB of storage when storing compressed mid-level features, whereas all the remaining methods use 3 GB of storage that corresponds to $20000$ raw images ($224\times 224$ uint8). Except for REMIND, these methods also store the current data corresponding to $100$ classes during each learning step. Hence, their total memory usage is much more than the replay memory budget (at least $7\times$ more). All methods use ResNet18~\cite{he2016deep}.


\section{Results}

\subsection{Evaluation on ImageNet-1K}

\begin{table}[t!]
  \caption{Comparison of CL methods evaluated on ImageNet-1K.
    $\#$P indicates total number of parameters in Millions. $\Omega$ refers to NetScore which is calculated based on top-5 final accuracy ($\%$), number of parameters, amount of memory and compute (updates). 
    Acc. is top-5 final accuracy ($\%$). Mem. is the total memory/storage used in Gigabytes. Comp. denotes compute in terms of total number of backprop updates in Millions. DER without pruning and DER with pruning are abbreviated as DER w/o P and DER w/ P, respectively.  Total $\#$P used by DER w/ P is unknown. All methods except End-to-End and Simple-DER follow same data ordering from~\cite{yan2021dynamically}. Best values are indicated in bold.
    \label{tab:inc}}
  \resizebox{\columnwidth}{!}{%
  \centering
     \begin{tabular}{lrrrrr }
     \hline
     Methods & $\#$P & Acc. & Mem. & Comp. & $\Omega$ \\
     \hline
     Offline & $11.68$ & $89.08$ & $192.89$ & $115.31$ & $27.53$ \\
     \hline
     iCaRL & $\mathbf{11.68}$ & $44.00$ & $22.32$ & $79.94$ & $5.62$ \\
     End-to-End & $\mathbf{11.68}$ & $52.29$ & $22.32$ & $93.26$ & $12.14$ \\
     BiC & $\mathbf{11.68}$ & $73.20$ & $22.32$ & $119.91$ & $24.97$ \\
     WA & $\mathbf{11.68}$ & $81.10$ & $22.32$ & $133.23$ & $28.80$ \\
     Simple-DER & $28.00$ & $80.76$ & $22.39$ & $213.17$ & $25.27$ \\
     DER w/o P & $116.89$ & $\mathbf{82.86}$ & $22.74$ & $213.17$ & $22.68$ \\
     DER w/ P & --- & $81.89$ & $22.32$ & $213.17$ & --- \\
     REMIND & $\mathbf{11.68}$ & $79.43$ & $\mathbf{2.05}$ & $\mathbf{58.78}$ & $\mathbf{35.99}$ \\ 
     \hline
    \end{tabular}}
\end{table}

Following the  common CL practice for ImageNet-1K~\cite{russakovsky2015imagenet}, each method was incrementally trained on 10 batches, where each batch has images from 100 classes that are only seen in that batch. The first batch serves as the base initialization phase, and the subsequent $9$ batches serve as the CL phase. We measure efficiency only for the CL phase. For all CL methods and the offline ResNet18 model, we compare the top-5 accuracy, number of parameters, amount of memory used, number of backpropagation updates used for training, and NetScore. 

Results are summarized in Table~\ref{tab:inc}, Fig.~\ref{fig:cost_acc}, and Fig.~\ref{fig:stratified-perf}. Simple-DER and End-to-End are abbreviated as Sim-DER and E2E respectively in Fig.~\ref{fig:cost_acc}.
In terms of NetScore, both REMIND and WA outperform the offline model. REMIND uses $10\times$ less memory and the least amount of compute. Shockingly, several methods use roughly as much or more compute than the offline model, including BiC, WA, and DER variants. While DER without pruning achieves the closest accuracy to the offline model, this comes at a cost of requiring $10\times$ more parameters and almost twice the compute as the offline ResNet18. Except for REMIND, all other systems require storing the entire batch in a storage.

\section{Discussion}

In this paper, we have argued that there is more to CL than catastrophic forgetting, and that the research community is ignoring critical factors e.g., model size, computational overhead, training time that are essential to address for CL to have real-world impact. We illustrated that multiple recent methods use as much or more compute than an offline learner, which is in opposition to one of the major goals of CL: efficient learning.

We believe CL can help to reduce the economic and environmental costs of deep learning, but this will only occur if CL algorithms are computationally efficient. Large DNNs require massive amounts of electricity to train, which greatly contribute to the growing amount of carbon emission worldwide. For example, the $176$-billion parameter language model BLOOM has been estimated to emit $50.5$ tonnes carbon in its life cycle~\cite{luccioni2022estimating}. In \cite{patterson2021carbon}, training GPT-3 was estimated to require as much energy as the annual consumption of $120$ U.S. homes. CL models can help address this issue and offer additional functionality, but only if they are computationally cheaper than periodic retraining of offline models as the dataset grows. 

Focusing on incremental class learning has led to many systems to be designed to only handle this scenario, which we consider an extreme edge case. They have bespoke features for this problem that are only appropriate if each batch contains unique classes, and their algorithms without major modifications break when this assumption does not hold. This assumption is invalid for almost all real-world applications, where we cannot make any assumptions about the distribution of new data. Ideally, a good CL system should learn from data in any order, including independent and identically distributed (iid) data or data that has temporal correlations. Of the methods we compared, only REMIND~\cite{hayes2020REMIND} has been designed to handle data with arbitrary orderings and to allow classes to be revisited.

We focused on CL methods that have been demonstrated to perform well on ImageNet-1K because it is much more closely aligned with real-world applications for DNNs than many of the tasks studied in the CL community. CL on ImageNet is just supervised learning, except we are progressively updating the DNN. In contrast, many CL systems are only evaluated on small-scale problems where they make additional assumptions beyond simply that the training data arrives incrementally.
For example,  many works still focus on tasks such as permuted MNIST and split-CIFAR100~\cite{chaudhry2018efficient, chaudhry2019tiny, kaushik2021understanding, mirzadeh2020understanding, titsias2019functional, pan2020continual, zenke2017continual, van2020brain}, 
which do not align with real-world supervised learning tasks where CL could be a drop-in alternative to periodically retraining from scratch. Many recent CL systems also assume task labels are available during training and/or evaluation~\cite{golkar2019continual, fernando2017pathnet, hung2019compacting, serra2018overcoming}, which is also not aligned with typical real-world applications of supervised learning.

For CL to have real-world impact, we make the following recommendations to the CL research community:
    \begin{enumerate}
        \item CL researchers need to focus on more than catastrophic forgetting. We argue that computational efficiency is the most critical factor, but other factors such as the number of additional parameters or memory used need to be taken into account. A CL system that uses more computation than an offline learner is hard to justify.
        \item Researchers should aim toward scaling up dataset sizes and the scope of the CL problems studied. While other areas of machine learning, e.g., generative methods, rapidly advanced such that reporting results on only toy datasets is unacceptable, much of the CL research community continues to focus on problems where CL is unnecessary. 
        \item Papers should clearly justify the CL paradigm they are studying and the limitations of their algorithm. Systems should be aligned toward real-world CL applications, unless trying to test a specific hypothesis regarding CL. For example, to enhance transparency in interpretation of experimental results a metric such as CLEVA~\cite{mundt2022clevacompass} could be used.
        \item Papers proposing new CL algorithms should report what an offline learner achieves and they should report the amount of compute needed to train the CL system relative to an offline learner. 
        \item CL systems should be tested robustly across multiple orders and should not  be designed only for extreme edge-cases, e.g., only being capable of incremental class learning. For real-world CL applications, we typically cannot make any strong assumptions about the distribution of the training data stream.
        \item Ideally, CL systems should be designed to be updated online, where if batches are used the system's robustness is assessed across multiple batch sizes. Many systems require extremely large batches to learn~\cite{yan2021dynamically, douillard2022dytox}, which may be acceptable for some applications but makes the system unacceptable for others, e.g., on-device learning. 
    \end{enumerate}

\section{Conclusion}
In this paper, we studied the efficiency of recent CL methods in terms of compute, memory, and accuracy. We found that some state-of-the-art techniques use more compute than the equivalent offline learner, which for many industrial applications makes them irrelevant even if they avoid catastrophic forgetting. We urge the research community to take factors beyond catastrophic forgetting into consideration. Systems must be tested to work across data orderings, in addition to being efficient in terms of compute, and for many applications, e.g., on-device learning, must also take into account the number of parameters and memory/storage that they use for training. We believe CL can play a critical role in reducing the global energy expenditure resulting from training DNNs, but this requires the community to align their work with the needs of industry.

\ifthenelse{\boolean{ack}}{
\paragraph{Acknowledgements.}
This work was supported in part by NSF awards \#1909696 and \#2047556.
}

{\small
\bibliographystyle{ieee_fullname}
\bibliography{egbib}
}

\end{document}